\documentclass[11pt]{article} 

\newcommand{\E}{\mathbb{E}}
\newcommand{\V}{\mathbb{V}}
\newcommand{\R}{\mathbb{R}}

\newcommand{\Z}{\mathbb{Z}}

\newcommand{\Prob}{\mathbb{P}}

\newcommand{\logtwo}{\log_{2}}
\newcommand{\KL}[2]{\mathcal{D}_{KL}\left(#1 \mid \mid #2 \right)} 
\newcommand{\SymKLexp}[2]{\frac{\KL{p}{q} + \KL{q}{p}}{2}}

\newcommand{\tp}[1]{{#1}^{\ensuremath{\mathsf{T}}}}  

\usepackage{float}
\usepackage[authoryear,sort&compress,longnamesfirst]{natbib}
\usepackage[margin=1.4in]{geometry}
\usepackage{amssymb, amsmath}
\usepackage{graphicx}
\usepackage{amsthm}
\usepackage{verbatim}
\usepackage[colorlinks,citecolor=blue,urlcolor=blue]{hyperref}
\usepackage{algorithm}
\usepackage[noend]{algorithmic}
\usepackage{tikz}
\usetikzlibrary{shapes,arrows}
\usepackage{fancyhdr}
\usepackage{url}
\usepackage{subfig}
\usepackage{setspace}

\newtheorem{theorem}{Theorem}[section]

\newtheorem{definition}[theorem]{Definition}

\graphicspath{{../images/}}
\DeclareGraphicsExtensions{.pdf}

\title{A Nonparametric Frequency Domain EM Algorithm \\ for Time Series Classification \\ with Applications to Spike Sorting and Macro-Economics}

\author{
Georg M.\ Goerg\thanks{\url{gmg@stat.cmu.edu}; \url{www.stat.cmu.edu/\textasciitilde gmg} } \\
Carnegie Mellon University, Department of Statistics
}

\begin{document}
\doublespacing
\maketitle

\begin{abstract}
I propose a frequency domain adaptation of the Expectation Maximization (EM) algorithm to group a family of time series in classes of similar dynamic structure. It does this by viewing the magnitude of the discrete Fourier transform (DFT) of each signal (or power spectrum) as a probability density/mass function (pdf/pmf) on the unit circle: signals with similar dynamics have similar pdfs; distinct patterns have distinct pdfs. An advantage of this approach is that it does not rely on any parametric form of the dynamic structure, but can be used for non-parametric, robust and model-free classification. This new method works for non-stationary signals of similar shape as well as stationary signals with similar auto-correlation structure. Applications to neural spike sorting (non-stationary) and pattern-recognition in socio-economic time series (stationary) demonstrate the usefulness and wide applicability of the proposed method.
\end{abstract}


\section{Introduction}
Classification of similar signals is a widespread task in signal processing, where similar can either mean similar shape (for non-stationary signals) or similar dynamics (for stationary\footnote{A sequence of random variables (RVs) $\lbrace x_t \rbrace_{t \in \Z}$  is stationary if i) $\E x_t = \mu < \infty$, ii) $\V x_t := \E (x_t - \mu)^2 < \infty$ and iii) the auto-covariance function $\gamma(k):=\E (x_t - \mu) (x_{t-k} - \mu)$ is independent of $t$.} signals). Non-stationary examples are recordings of brain activity (see Section \ref{sec:spike_sorting}) or speech signals; stationary signals can be found in many economic or physical time series. In both cases, researchers want to detect similar dynamics:
\begin{description}
\item Neuro-scientists study the signal shape sent by neurons in order to understand how fast neurons send information across the brain. As a recording can contain signals from many different neurons, it is necessary to cluster them into signals of similar shape \citep{Quiroga04}, which were presumably sent by the same neuron.
\item In economics and public policy one is often interested in similar dynamics of the market/society to characterize, for example, how fast a country recovers from a recession, and how it compares to other countries in the region; or which countries have similar dynamics in their labor market.
\end{description}

Formally, let $\mathcal{X} = \lbrace \mathbf{x}_{1,t}, \ldots, \mathbf{x}_{N,t} \rbrace$ be a family of sequential observations from a dynamical system $\mathcal{S}$, where $\mathbf{x}_{i,t} = (x_{i,1}, \ldots, x_{i,T})$ is the individual time series of entity $i$. For example, $\mathcal{S}$ can be a particular area in the brain or the economic rules in the labor market. Here we consider systems which can be naturally divided into $K$ homogeneous sub-systems $\mathcal{S}_1, \ldots, \mathcal{S}_K$, each one with its own characteristic dynamics. In the neurology context these sub-systems $\mathcal{S}_k$ represent different neurons sending a signal; in economics  $\mathcal{S}_k$ could correspond to different dynamics in the market, e.g.\ countries that recover fast from a recession ($\mathcal{S}_1$) versus countries that need more time to catch up again to global economy ($\mathcal{S}_2$).

Many clustering and dimension reduction techniques such as principal component analysis (PCA) \citep{Jolliffe02_PCAbook} focus on the mean and/or variance to make a reduction/classification in similar blocks of data. Yet these two statistics are irrelevant for the correlation over time; \citet{Keogh03_MeaninglessTSclustering} even claim that time series clustering is entirely meaningless. \citet{Simon05_UnfoldingTScluster} show that time series clustering is not meaningless \emph{per se}, but that the similarity measure must be chosen carefully. They embed each time series in a higher dimensional space of lagged variables, $x_t \rightarrow \tp{\left(x_{t-\tau_1}, x_{t-\tau_2} \ldots, x_{t-\tau_s} \right)} \in \R^{s}$, $0 \leq \tau_1 < \tau_2 < \ldots < \tau_s < T$, such that signals with different dynamics can be easily distinguished in the higher dimensional $\R^s$. This method works particularly well for long time series even with non-linear dynamics. If only few observations per series are available ($T \approx 100$ or even only $50$), then time-embeddings are extremely sparse in $\R^{s}$ and thus clustering becomes impractical.

For few observations per series it can be useful to first fit a parametric model $\mathcal{M}_{\theta_j}$ to every series $\mathbf{x}_{j,t}$, and then cluster in the lower-dimensional parameter space $\lbrace \widehat{\theta}_{1}, \ldots, \widehat{\theta}_{n} \rbrace$. For example, for the broad class of auto-regressive integrated moving average (ARIMA) models several approaches have been studied: \citet{Dhiral01_ClusteringARIMA} cluster ARIMA models based on the distance between their estimated coefficients; \citet{Piccolo90_ARIMAdistance} uses the Euclidean distance between their auto-regressive extensions as a metric on the invertible ARIMA model space; \citet{Maharaj00} present a hypothesis test to distinguish between two - not necessarily independent - stationary time series by comparing auto-regressive fits to the data. See \citet{Liao05} for a detailed survey. Although this works well for small $T$, it suffers from a model selection bias: if we pick the wrong model for just some of the series, then the clustering cannot be accurate anymore. Furthermore, if the models are not nested in some sense, then it is hard to compare the parameters of $\mathbf{x}_{j,t}$ to those of $\mathbf{x}_{i,t}$.\\

Here I propose a novel approach to clustering similar dynamics using frequency domain properties of the signals, which avoids the model selection bias and at the same time works even with few observations. Existing frequency domain classification methods are mostly based on defining a metric on the spectrum and then using a clustering algorithm based on the so-obtained distance (or similarity) matrix. \citet{CaiadoNunoPena06} use hierarchical clustering algorithm on the Euclidean distance between the log-spectra; \citet{Savvides08} use a distance measure on cepstral coefficients obtained from the log-spectra. The method proposed here differs from existing techniques as it treats the magnitude of the discrete Fourier transform (DFT) of signal $\mathbf{x}_{j,t}$ as a probability mass function (pmf) on the unit circle and thus leads to a natural classification by an adaptation of the well-known Expectation Maximization (EM) algorithm \citep{Dempster77_EM, Bishop07_ML_book}. Section \ref{sec:EM_spectral_density} describes a non-parametric version which avoids the model selection bias, but it can also be easily adapted to a parametric framework, e.g.\ to cluster time series within the ARIMA model class.

\subsection{Similar dynamics in socio-economic time series}
In macro-economics and public policy researchers are often interested in comparing economies/societies with each other. For example, annual unemployment rates over the course of several decades can show law changes or adaptations of economic interdependencies within a country as well as with the rest of the world. 

Here I will consider the annual per-capita income growth rate of the ``lower 48'' in the US from $1958$ to $2008$ compared to the overall US growth rate
\begin{equation}
g_{j,t} := r_{j,t} - r_{US,t}, \quad j \in \lbrace \text{Alabama, $\ldots$, Wyoming} \rbrace,
\end{equation}
where $r_{j,t}$ is the annual growth rate of region $j$ (see Appendix \ref{sec:Data} for details). Clustering states according to similar economic dynamics can help to decide where to provide economic support to overcome a recession faster. For example, if certain states do not show any important dynamics on a 7-8 year period - which is typically considered the ``business cycle'' \citep{Halletetal08_EuropeanBusinessCycle, Iacobucci03_spectralanalysiseconomics} - then it might be more useful for to invest available money in those states that are heavily affect by these global economy swings.

This dataset has also been analyzed in \citet{Dhiral01_ClusteringARIMA}, who fit auto-regressive models of order $1$ ($AR(1)$) to the non-adjusted growth rates $r_{j,t}$ for pre-selected $25$ states, and then cluster them based on the different fits. Although this procedure gives useful results, it is very unlikely that different dynamics for each of the $48$ states only manifest themselves in a different $AR(1)$ coefficient. In particular, simple $AR(1)$ models cannot capture the business cycle dynamics which are clearly visible in the power spectra of the growth rates (even in the adjusted rates) - see Section \ref{sec:marco_econ_results}, Fig.\ \ref{fig:EM_spikes_periodogram}.

The non-parametric EM algorithm introduced in Section \ref{sec:EM_spectral_density} does not face this model selection bias, but can capture different cyclic components in all $48$ time series.

\subsection{Neuron identification - ``spike sorting''}
\label{sec:spike_sorting}
\begin{figure}[!t]
\centering
\makebox{\includegraphics[width=0.8\textwidth]{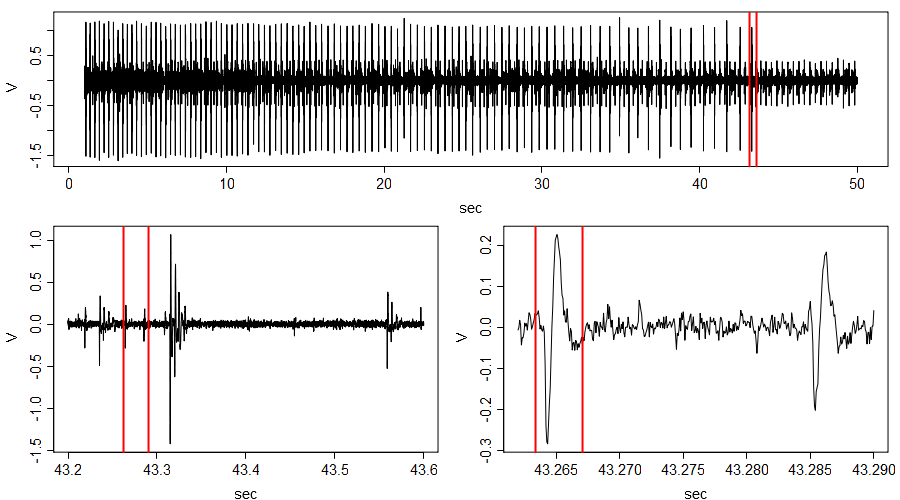}}
\caption{\label{fig:data_recording_40_50} Brain signal recording: (top) entire $49$ seconds; (bottom left) zoom to $[43.2,43.6]$ - the transition between extremely large spikes ($\sim 43.32$) and smaller spikes ($\sim 43.57$); (bottom right) zoom to $[43.26, 43.29]$ where two spikes become visible}
\end{figure}

The human brain can be seen as a big information-processing and -storing unit. For example, the information we get from watching our environment must be carried from the eye to the visual cortex. As the visual cortex resides in the back of the brain neurons have to transmit information from the front to the back of the head, just for us to being able to make sense of what we see; set aside the neurons involved in executing our reaction to what we see. Every time a neuron transmits information it emits an electrochemical signal, which can be measured by an electrode put in the brain area of interest. Figure \ref{fig:data_recording_40_50} (top) shows a recorded signal $y_t$ with $73,500$ observations.\footnote{For a detailed description see Appendix \ref{sec:Data}.}.

On a macro-level these measurements help to identify active areas of the brain which are involved in performing a particular task; e.g.\ for visual tasks the back of the brain shows up as an active area. Micro-level properties of neuron activity are also important: for example \citet{KassVentura01} analyze how fast neurons can send information  - this is characterized by a neuron's ``firing rate''. To do this non-trivial task, however, one implicitly makes an important assumption: it is known which neuron sent which signal. Figure \ref{fig:data_recording_40_50} clearly shows that micro-electrodes cannot single out one neuron, but record a concatenation - and sometimes a superposition - of an unknown number of neurons $n_1, \ldots, n_K$ transmitting information plus a lot of background noise. Hence to successfully analyze the firing rates, it is necessary to
\begin{enumerate}
\item[i)] distinguish actual spikes from background noise, and
\item[ii)] identify and assign each signal to one particular neuron $n_k$, $k = 1, \ldots, K$ where the number of neurons $K$ is unknown: an electrode records as many neurons as there are in its local neighborhood.
\end{enumerate}

Part i) constitutes one of the core problems in signal processing \citep{ZhengmingXiaojun00, DaviesJames07}. Consequently there is an immense literature on signal/noise separation, especially in audio and speech processing \citep{Jangetal03_SingleChannel, Barry05}. For sub-problem ii) we can classify the observed spikes into classes of similarly shaped wave-forms. If these shapes actually correspond to one sole neuron $n_k$ or still to a collection of neurons, depends on whether each neuron has a unique wave form or not. Only if there exists such a one-to-one relation, we can determine the firing rates of each single neuron. Biochemical and physiological findings suggest that each neuron has its own unique wave-form, which can only vary slightly based on the state of the neuron. Thus it should be possible to classify neuron activity according to the form of the signal - the ``spike''. This classification task is commonly known as ``spike sorting'' \citep{Lewicki98,Kim06,Natakanietal01}.

A common and simple approach is performing PCA on the spikes, and then cluster the signals according to the PCA coefficients \citep{Wood04_Automaticspike}. Although generally there are far more spikes than observations per spike ($N \gg T$), still the first $2$-$3$ eigen-vectors of the low-rank correlation matrix capture most of the variation in the data. However, since PCA selects sources by the direction of maximum variance, it will classify low power firings from the same neuron as different neurons.

The frequency domain classification algorithm introduced here, builds on the relation between the shape of the signal and its Fourier coefficients. Similar shapes have similar Fourier coefficients and thus clustering in the frequency domain should reveal these sub-classes.

\section{``Spike sorting'' in the time domain}
Let $\mathcal{N}$ be the set of all neurons and assume that each neuron $n_i \in \mathcal{N}$ has a unique characteristic spike $S_i(t) \in  \mathcal{C}[a,b]$, where $ \mathcal{C}[a,b]$ is the set of all continuous functions 
 on $[a,b]$. The spike is unique in the sense that $n_i = n_j$ if and only if $S_i(t) = S_j(t)$, or put in words if we see two different spikes, then we know that two different neurons were active and vice versa.

The micro-electrode only records a subset of neurons $n_k$, $k = 1, \ldots, K$, where $K$ is unknown. In Section \ref{sec:spike_detection} I use a slowness measure to distinguish between signal and noise, and in Section \ref{sec:features} I fit a Gaussian mixture model (GMM) to the slowness to detect different neurons, based on the assumption that a every different spike shape has its characteristic slowness.

\subsection{Spike detection}
\label{sec:spike_detection}
Given the recorded signal $y_t$ it is necessary to extract windows of size $T$ containing a spike.\footnote{Since these extracted windows containing a spike will later on be used as the $N$ time series $\lbrace \mathbf{x}_{1,t}, \ldots, \mathbf{x}_{N,t} \rbrace$ of length $T$ to the classification algorithm, I also use $T$ here to denote the window size.} These signals $s_{j, t}$ of length $T$ represent the family of sequential observations $\mathcal{X} = \lbrace s_{1, t}, \ldots, s_{N, t} \rbrace \in \R^{T \times N}$, where $N$ is the number of detected spikes. As the entire micro-electrode recording is much longer than the length of one single spike there are far more extracted spikes than number of observations per signal ($N \gg T$). Since the electrode only records signals in its local neighborhood, we can also expect a small number of sub-systems (neurons) $\mathcal{S}_1, \ldots, \mathcal{S}_K$ of similar shape ($K \ll N$). The size of the window must match the length of a typical spike: the lower right panel of Fig.\ \ref{fig:data_recording_40_50} suggests that a typical spike lasts for about $0.0035$ seconds $\approx$ $55$ time steps (vertical red lines). Thus for the rest of this section I set $T = 55$.

Since we do not know a-priori where a spike occurs we need a rule that tells us where to look for it. Whereas characterizing spikes visually is easy, designing a quantitative automated rule that can describe spikes is much more difficult. A common approach \citep{Quiroga04} is to set a threshold value $tol$ and a spike is detected if the signal exceeds this threshold. This threshold rule will not only be very sensitive to outliers, but also bias the selected spikes in favor of spikes with large variance (power). Furthermore neurons sometimes fire with lower power than usual, and thus may not exceed such a threshold. Although missing these spikes would not affect the spike sorting algorithm, it will underestimate the firing rate of neuron $n_k$. 

Here I characterize ``non-spikes'', i.e.\ noise, in a way that detects spikes according to properties of the entire signal, not of one single observations (such as the threshold rule). One way to characterize noise is that it is moving much faster than any spike - whatever such a spike may look like. \citet{Berkes05} introduced a measure of slowness for a signal $x_t$, defined as the variance of the differenced, unit-variance signal
\begin{equation}
\label{eq:slowness}
\Delta(x_t) = \V \left( x_t - x_{t-1} \right), \quad \V x_t = 1.
\end{equation}
For  an independent identically distributed (iid) signal $\varepsilon_t$ the slowness satisfies $\Delta\left(\varepsilon_t\right) = 2$. On the other hand, if $x_t \rightarrow const$ then $\Delta(x_t) \rightarrow 0$. Therefore, the larger $\Delta(x_t)$, the faster $x_t$. 

\begin{figure}[!t]
\centering
\subfloat[(top): (red) Simulation of $\Delta \left(\varepsilon_t\right)$ with $10,000$ replications for iid $\lbrace \varepsilon_t \rbrace_{t=1}^{T=55}$; (blue) empirical slowness of the rolling window over the data $y_t$. \newline (bottom): zoom into $(0, 0.4)$ with the boundary between spikes ($ < 0.25$) and noise ($> 0.25$).]{\includegraphics[width=.45\textwidth]{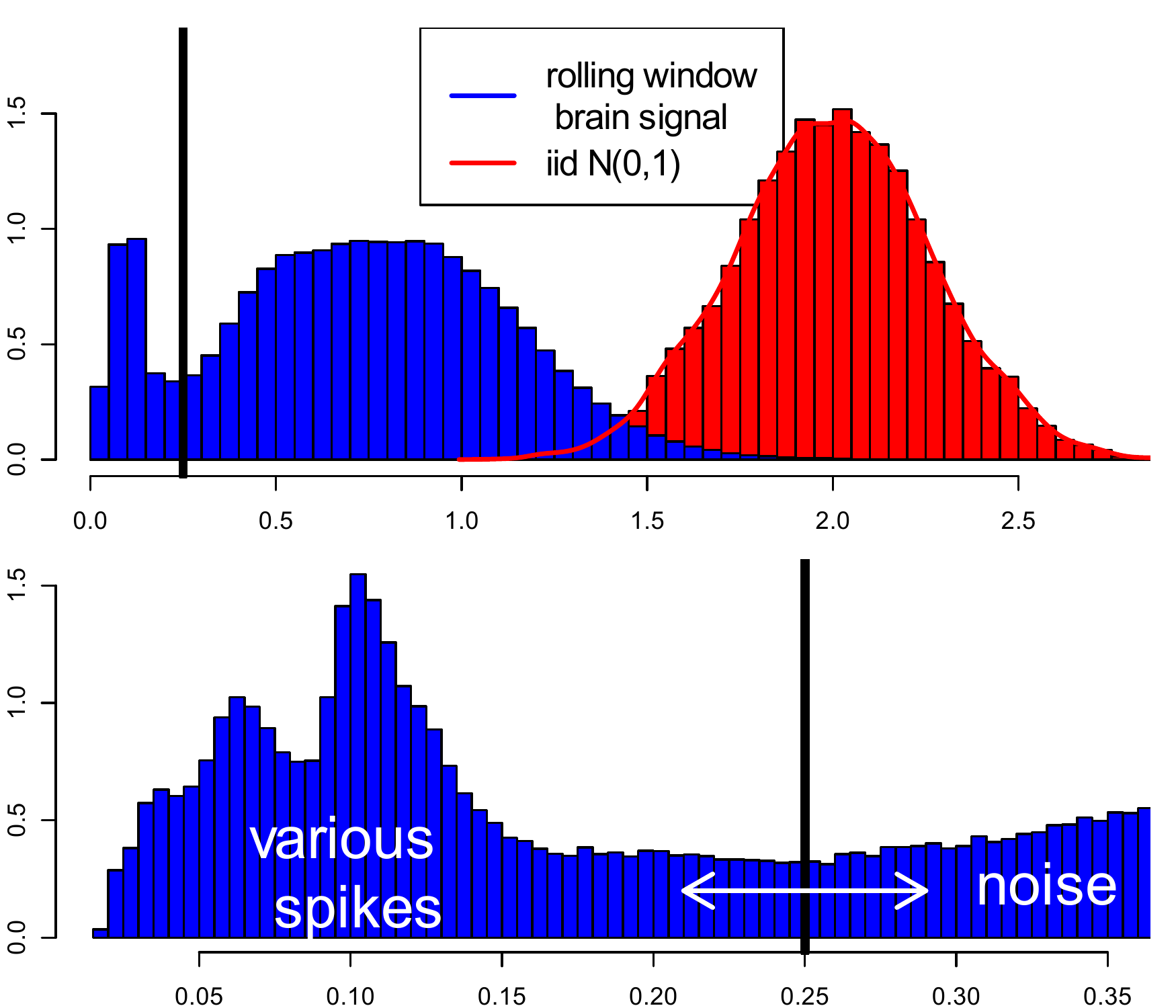}\label{fig:MonteCarlo_slowness} }
\hspace{1cm}
\subfloat[(top) detected spikes; (below) their $\log$ slowness and a Gaussian mixture fit with $6$ components (chosen according to BIC score)]{\includegraphics[width=.45\textwidth]{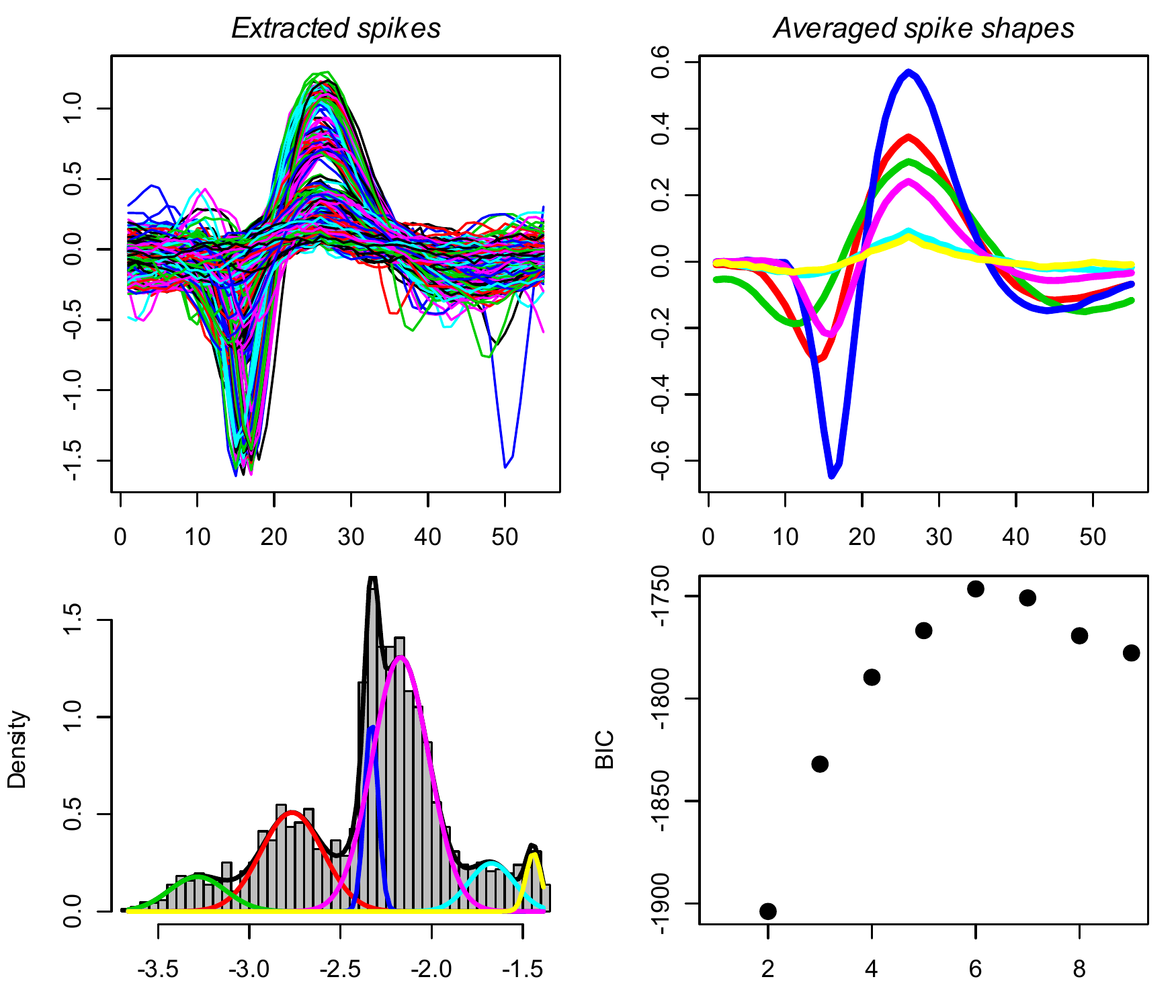}\label{fig:extracted_spikes} }
\caption{How slowly do we think? - the slowness of brain recordings.}
\label{fig:Flatness}
\end{figure}

Computing the slowness of the signal in a sliding window over $y_t$ reveals noisy parts (fast) and - complementary - the spikes (slow). The red (right) histogram in Fig.\ \ref{fig:MonteCarlo_slowness} shows simulated $\Delta\left( \varepsilon_t \right)$, where $\varepsilon_t \stackrel{iid}{\sim} \mathcal{N}(0,1)$ with $t = 1, \ldots, T = 55$ for $N = 10,000$ replications. Clearly, the central limit theorem (CLT) comes into play and the simulated values are centered around their true slowness $\Delta\left(\varepsilon_t\right) = 2$.

However, there is no obvious reason to assume that the brain background noise in the neighborhood of the micro-electrode is necessarily iid. In fact, the empirical slowness (blue histogram) of the sliding windows is substantially smaller than $2$, showing that brain background noise is not iid.\footnote{Since $\Delta\left( \varepsilon_t \right) = \Delta\left( const \cdot \varepsilon_t \right)$ by definition ($\V x_t \equiv 1$ in \eqref{eq:slowness}), the lower slowness for the brain signal is not due to a lower variance white noise sequence, but indeed a manifestation of some dependence in the data.}
But even though we do not know how slow it is, we know - and can clearly see in Fig.\ \ref{fig:MonteCarlo_slowness} (bottom) - that noise moves much faster than any of the spikes: $\Delta(\text{background noise}) \gg \Delta\left( \text{any spike} \right)$. Hence, we can learn the boundary value that distinguishes noise and spikes from the data. At this stage we are only concerned with separating spikes from noise, thus we can choose a conservative value for the boundary. If it turns out that this still includes too much noise, then a clustering algorithm will put these falsely extracted ``spikes'' in a \emph{noise} class. On the other hand, a too small boundary will miss spikes and thus bias the analysis of firing rates towards larger firing intervals. The lower panel of Figure \ref{fig:MonteCarlo_slowness} suggests that $tol = 0.25$ provides a good separation between noise on the right and spikes on the left.

This rolling window approach gives the so called on-set times (the moment a neuron fires and the spike lasts for $T = 55$ units of time), which are then used to extract possible spikes $s_{j,t}$ from $y_t$. An additional alignment step takes place to avoid slight misplacements of the onset times; following the spike sorting literature, this was done by identifying the maximum of each spike and adjust the window such that all signals have their maximum at the same position.

Figure \ref{fig:MonteCarlo_slowness} shows $n = 1,747$ extracted signals of length $T = 55$ obtained by applying the slowness measures on a sliding window using $tol = 0.25$. The rolling window spike detection could not exclude noise completely, so in the upper left panel of Fig.\ \ref{fig:extracted_spikes} one can visually distinguish $2$-$3$ spikes and some noise. Even though the low-variance signals seem to be noise, they could also be just low-power spikes. Since the slowness measure is invariant to scaling, it does not falsely ignore low power signals. 

Before applying a standard classification algorithm on the extracted signals in Section \ref{sec:Results}, I first describe the main contribution of this work.
\section{Non-parametric frequency domain EM algorithm}
\label{sec:EM_spectral_density}

The fundamental idea of this novel approach to clustering dynamic structures in time series is to identifying each time series with the distribution it induces on the unit circle - and thus on the interval $[-\pi, \pi]$ - by its Fourier transform. These distributions can then be used in a mixture density setting and an adaptation of the EM algorithm yields the classification algorithm.

\begin{definition}[Spectral Density]
The spectral density of a stationary, zero-mean time series $x_t$ with auto-covariance function $\gamma_{x}(\ell) = \E x_t x_{t-\ell}$ is defined as
\begin{equation}
\label{eq:spectral_density}
f_x(\lambda) := \frac{1}{2 \pi} \sum_{\ell=-\infty}^{\infty} \gamma_{x}(\ell) e^{i \lambda \ell}, \quad \lambda \in [-\pi, \pi],
\end{equation}
where the limit is understood point-wise if $\lbrace \gamma_{x}(\ell) \rbrace_{\ell=-\infty}^{\infty}$ is absolutely summable, and in the mean-square sense if $\lbrace \gamma_{x}(\ell) \rbrace_{\ell=-\infty}^{\infty}$ is square summable.
\end{definition}
For real valued processes $\gamma_{x}(\ell) = \gamma(-\ell)$, thus $f_x(\lambda) \geq 0$ for all $\lambda$. Furthermore,  
\begin{equation}
\label{eq:spectral_decomposition}
\int_{-\pi}^{\pi} f_x(\lambda) = \sigma_x^2,
\end{equation}
since $\int_{-\pi}^{\pi} e^{i \lambda \ell} d \lambda = 0$ if $\ell \neq 0$ and $\gamma_x(0) = \sigma_x^2$. Equivalence \eqref{eq:spectral_decomposition} is also known as the spectral decomposition of the variance of a time series. Hence, the spectral density is a non-negative function on the interval $[-\pi,\pi]$ and peaks at $\lambda_0$ indicate that this frequency is important for the overall variance of the process, since those peaks contribute a lot to the integral in \eqref{eq:spectral_decomposition}.

An estimate of the spectral density is the power spectrum or \emph{periodogram}.
\begin{definition}[Periodogram]
The periodogram (or power spectrum) of $x_t$ is defined as
\begin{equation}
\label{eq:periodogram}
I_{T,x}(\omega_j) := \left| X(\omega_j) \right|^2 = \Big| \frac{1}{\sqrt{T}} \sum_{k=0}^{T-1} x_t e^{-2 \pi i \omega_j t}\Big|^2, \quad \omega_j= j/T, \quad j=0,1,\ldots, T-1
\end{equation}
where $\omega_j$ are the Fourier frequencies (scaled by $2 \pi$ for easier interpretation). 
\end{definition}

For large $T$ a frequent model for $I_{T,x}(\omega_j)$ is \citep[see][]{BrockwellDavis91}
\begin{align}
I_{T,x}(\omega_j) = 
\begin{cases} 
\chi_1^2, & \text{if $j = 0$ or $T/2$,}
\\
f_x(\omega_j) \, \eta, & \text{otherwise.}
\end{cases}
\end{align}
where $\eta$ is a standard (rate $=$ 1) exponential RV. At each frequency $\omega_j$ (except $0$ and $\pi$) the periodogram is an exponential RV with rate parameter equal to the true spectral density $f_x(\omega_j)$. Therefore $I_{T,x}(\omega_j)$ is asymptotically unbiased ($\E I_{T,x}(\omega_j) =  f_x(\omega_j)$), but not consistent ($\V I_{T,x}(\omega_j) =  f_x(\omega_j)^2  \stackrel{T \rightarrow \infty}{\nrightarrow} 0)$. This is especially harmful for large values of the true spectral density, as they exactly correspond to those frequencies which are particularly important for the overall variation.\\

There are two main ways to reduce the variance of the raw periodogram. If only one time series is available, then one can reduce the variance of $I_{T,x}(\omega_j)$ by smoothing over neighboring frequencies  \citep{OppenheimSchafer89_Discrete-TimeSignalProcessing} - just as in non-parametric density estimation. This works well for series with many observations, but for small samples such as the neuron spikes or many economic time series averaging over neighboring frequencies is not a practical option as it quickly introduces too much bias at each $\omega_j$. 

For $M_k$ independent time series $\lbrace x_{m,t} \rbrace_{m=1}^{M_k} $ of the same type (all from sub-system $\mathcal{S}_k$) a variance-reduced estimate of the true $f_{\mathcal{S}_k}(\lambda)$ can be obtained by averaging over all $M_k$ periodograms at each frequency
\begin{equation}
\label{eq:M_periodograms}
\widehat{f}_{\mathcal{S}_k}(\lambda) \mid_{\lambda = \omega_j} = \frac{1}{M_k} \sum_{m = 1}^{M_k} I_{T, x_{m,t}}(\omega_j), \quad j=0, \ldots, T-1.
\end{equation}
Since by assumption all $x_{m,t} \in \mathcal{S}_k$ have the same dynamic structure, $\widehat{f}_{\mathcal{S}_k}(\lambda)$ is also a good estimate of $f_{x_{m,t}}(\lambda)$ for all $m = 1, \ldots, M_k$.

If the sub-series $x_{m,t}$ are far enough apart in a signal $y_t$, then periodograms $I_{T, x_{m,t}}(\omega_j)$ can be considered as independent estimates of the same underlying true spectral density $f_{\mathcal{S}_k}(\lambda)$. Thus \eqref{eq:M_periodograms} is still unbiased but has a much lower variance
\begin{equation}
\E \widehat{f}_{\mathcal{S}_k}(\omega_j) = f_{\mathcal{S}_k}(\omega_j), \quad \V \widehat{f}_{\mathcal{S}_k}(\omega_j) = \frac{ f^2_{\mathcal{S}_k}(\omega_j)}{M_k}.
\end{equation}

\subsection{From spectral density estimation to the EM algorithm}
Equation \eqref{eq:M_periodogram} looks very similar to the M step of an EM algorithm \citep{McLachlanetal08_EM_book}. By averaging over periodograms in \eqref{eq:M_periodograms} we assume we \emph{know} that series $\mathbf{x}_{i,t}$ came from system $\mathcal{S}_k$. This can be a reasonable assumption when repeatedly measuring time series in controlled physical experiments. In many applications, however, it is not \emph{known} where the signal came from. Thus I introduce a non-parametric frequency domain EM algorithm to classify time series. As a general idea this shift from averaging periodograms deterministically to probabilistically is analogous to the shift from hard-thresholding in k-means to soft-thresholding in the EM algorithm.\\

Formally, let $\mathbf{z}_{i}$ be a $K$-dimensional vector indicating from which system series $\mathbf{x}_{i,t}$ comes from; i.e.\ $z_{i k} = 1$ if $\mathbf{x}_{i,t}$ is from $\mathcal{S}_k$, $0$ otherwise. By averaging over periodograms as in \eqref{eq:M_periodograms}, $\mathbf{z}_i$ is treated as a deterministic, known variable. Thus \eqref{eq:M_periodograms} can be rewritten to
\begin{align}
\widehat{f}_{\mathcal{S}_k}(\lambda) \mid_{\lambda = \omega_j}  &= \frac{1}{M_k} \sum_{m = 1}^{M_k} I_{x_{m,t}}(\omega_j) \\
\label{eq:M_periodogram_z}
&= \frac{1}{\sum_{i = 1}^{N} z_{ik} }\sum_{i = 1}^{N} z_{ik} I_{\mathbf{x}_{i,t}}(\omega_j), \quad j=0, \ldots, T-1.
\end{align}

For the EM framework we treat $\mathbf{z}_i$ as random variable with marginal distribution $\Prob \left( z_{ik} = 1 \right) = \pi_k$, also commonly referred to as \emph{mixing weights}. Rather than weighing periodograms with binary weights in \eqref{eq:M_periodogram_z}, the non-parametric frequency domain EM estimator for $\widehat{f}_{\mathcal{S}_k}(\lambda)$ weighs the periodogram of series $\mathbf{x}_{i,t}$ with the probability of coming from system $\mathcal{S}_k$, that is 
\begin{align}
\label{eq:gamma_periodogram}
\widehat{f}_{\mathcal{S}_k}(\omega_j) &= \frac{1}{N_k}\sum_{i = 1}^{N} \gamma_{ik} I_{\mathbf{x}_{i,t}}(\omega_j),
\end{align}
where
\begin{equation}
\label{eq:gamma_ik}
\gamma_{ik} := \Prob \left( z_{ik} = 1 \mid \mathbf{x}_{i,t} \right),
\end{equation}
and $N_k = \sum_{i = 1}^{N} \gamma_{ik}$ is the effective number of time series from sub-system $\mathcal{S}_k$. As a by-result this new method also gives improved spectral density estimates.\\

For the frequency domain EM algorithm we treat the spectral density/periodogram of $\mathbf{x}_{i,t}$ as a pdf/pmf on the on the Fourier frequencies $\mathbf{\lambda}_i = \left( \lambda_{i,0}, \ldots, \lambda_{i,T-1} \right)$. Thus we compute \eqref{eq:gamma_ik} by the probability that ``model'' density $f_{\mathcal{S}_k}(\lambda)$ assigns to the empirical distribution function (edf) of the Fourier frequencies of $\mathbf{x}_{i,t}$ (= periodogram of $\mathbf{x}_{i,t}$), i.e.\
\begin{align}
\Prob\left( z_{ik} = 1 \mid \mathbf{x}_{i,t} \right) := \Prob\left( I_{\mathbf{x}_{i,t}}(\lambda) \text{ from } f_{\mathcal{S}_k}(\lambda) \right)
\end{align}
As we do not observe the Fourier frequencies $\mathbf{\lambda}_i$ we cannot compute likelihoods and probabilities such as $ \Prob\left( I_{\mathbf{x}_{i,t}}(\lambda) \text{ from } f_{\mathcal{S}_k}(\lambda) \right)$ directly. However, eq.\ \eqref{eq:loglik_equals_KL_ent} in the Appendix \ref{sec:KL_MLE} shows how to compute the log-likelihood of $\mathbf{\lambda}_i$ as a linear combination of the Kullback-Leibler (KL) divergence between $I_{\mathbf{x}_{i,t}}(\lambda)$ and $f_{\mathcal{S}_k}(\lambda)$, and the entropy of $I_{\mathbf{x}_{i,t}}(\lambda)$. 

Thus the EM algorithm can be implemented as follows:
\begin{enumerate}
\setcounter{enumi}{-1}
\item Initialization: set $\tau = 0$ and randomly assign $\mathbf{x}_{i,t}$ to one of the $K$ sub-systems; set class probabilities $\gamma_{ik}^{(\tau)} := 1$ if $\mathbf{x}_{i,t} \in \mathcal{S}_k$; $0$ otherwise. Compute effective number of time series per cluster $N_k^{(\tau)} = \sum_{i=1}^{N} \gamma_{ik}^{(\tau)}$ and estimate mixing weights by $\widehat{\pi}_k^{(\tau)} = \frac{N_k^{(\tau)}}{N}$.

\item \label{item:start_iteration} Estimate $f_{\mathcal{S}_k}(\lambda)$ by a weighted average of the periodograms of $\mathbf{x}_{i,t}$:
\begin{equation}
\label{eq:est_f_j2}
\widehat{f}_{\mathcal{S}_k}^{(\tau)}(\omega_{j}) = \frac{1}{N_k^{(\tau)}} \sum_{i=1}^{N} \gamma_{i{k}}^{(\tau)} I_{\mathbf{x}_{i,t}}(\omega_{j}), \text{ for each } k =1, \ldots, K.
\end{equation}
This gives $K$ spectral densities $\lbrace \widehat{f}_{\mathcal{S}_1}^{(\tau)}, \ldots, \widehat{f}_{\mathcal{S}_K}^{(\tau)} \rbrace =: \mathcal{F}^{(\tau)}$ at iteration $\tau$. Note that for each $k$, $\E \widehat{f}_{\mathcal{S}_k}^{(\tau)}(\omega_{j}) = f_{\mathcal{S}_k}(\omega_{j})$ and $\V \widehat{f}_{\mathcal{S}_k}^{(\tau)}(\omega_{j})\approx \frac{ f_{\mathcal{S}_k}(\omega_{j}) }{N_k} \ll f_{\mathcal{S}_k}(\omega_{j})$.

\item \label{item:end_iteration} Compute KL divergence between each $I_{\mathbf{x}_{i,t}}(\omega_{j})$ and all $\widehat{f}_{\mathcal{S}_k}^{(\tau)} \in \mathcal{F}^{(\tau)}$:
\begin{equation}
\label{eq:KL_div_periodograms}
\KL{I_{\mathbf{x}_{i,t}}}{\widehat{f}_{\mathcal{S}_k}^{(\tau)}} = \sum_{j=0}^{T-1} I_{\mathbf{x}_{i,t}}(\omega_{j}) \log \frac{I_{\mathbf{x}_{i,t}}(\omega_{j})}{\widehat{f}_{\mathcal{S}_k}^{(\tau)}(\omega_{j})}, \quad \forall i, \forall k,
\end{equation}
and update conditional probability that series $\mathbf{x}_{i,t}$ comes from system $\mathcal{S}_k$
\begin{align}
\gamma_{ik}^{(\tau+1)} &= 
\Prob\left( I_{\mathbf{x}_{i,t}}(\lambda) \text{ from } \widehat{f}_{\mathcal{S}_k}^{(\tau)} \right) \quad \forall i, \forall k
\end{align}
using \eqref{eq:KL_div_periodograms} and \eqref{eq:probability_from_loglik}. Update mixing weights
\begin{equation}
\widehat{\pi}_k^{(\tau+1)} = \frac{N_k^{(\tau+1)}}{N} \text{, where } N_k^{(\tau+1)} = \sum_{i=1}^{N} \gamma_{ik}^{(\tau+1)}.
\end{equation}

Set $\tau = \tau +1$.
\item Repeat steps \ref{item:start_iteration} and \ref{item:end_iteration} until convergence of the overall spectral likelihood
\begin{equation}
\label{eq:loglik_total}
\ell(\mathcal{S}_1, \ldots, \mathcal{S}_K; \pi_1, \ldots, \pi_K \mid \mathbf{x}_{1,t}, \ldots  \mathbf{x}_{N,t}) = \sum_{i=1}^{N} \log \left( \sum_{k=1}^{K} \widehat{\pi}_k e^{\ell \left( I_{\mathbf{x}_{i,t}}(\omega) \mid \widehat{f}_{\mathcal{S}_k}(\omega) \right) } \right).
\end{equation}
\end{enumerate}

Since for unit-variance input $x_t$ the spectral density/periodogram are well-defined continuous/discrete probability distributions, this EM algorithm can be applied to both stationary as well as non-stationary signals: in the first case, the spectral density $f_x(\lambda)$ exists as a non-negative, square integrable function and a large part of the time series and econometrics literature is devoted to the spectral analysis of stationary time series \citep{Iacobucci03_spectralanalysiseconomics, Priestly81_SpectralAnalysis,Mathias04_Irregularspaced}; in the second case, the periodogram \eqref{eq:periodogram}, viewed as a purely data-driven method, represents a valid discrete pmf on $\lbrace \omega_j \rbrace_{j=0}^{T-1}$.

Since frequency domain analysis plays a very prominent and successful role in statistics, time series analysis, and signal processing, this frequency domain EM algorithm to detect similar dynamics or shape can be easily implemented and applied to a great variety of problems where the data has a spectral representation. For example, the method can be used for image clustering ($2D$ Fourier transform) as well as classification of a family of positive semi-definite random matrices $\lbrace A_i \rbrace_{i=1}^{N}$, $A_i \in \R^{T \times T}$ considering their normalized eigenvalues $\lbrace \lbrace \tilde{\lambda}_{j} \rbrace_{j=1}^{T} \rbrace_i$ as a discrete distribution on $j = 1, \ldots, T$.\\

It must be noted though that it comes with all the pros and cons of the basic EM algorithm (never decreasing likelihood, but possibly local optima). For a detailed account of convergence results and many other properties see \citet{McLachlanetal08_EM_book} or \citet{Bishop07_ML_book}.

\subsection{Choosing the number of clusters}
So far the number of clusters was fixed a-priori and the algorithm gives the (locally) best $K$-cluster solution. However, since this number is rarely known in practice, we must have a rule to select a good $K$. In some cases there is a ``true'' $K$. For example, the micro-electrode in the brain recorded a certain number of neurons. Thus there is an underlying truth which we try to estimate. In other cases, such as the economic time series example, there may not be a true number of sub-systems but choosing the number of clusters is based on convenience and ease of interpretation. One may choose only two clusters to show vastly contrary situations, and then compare this to a more refined structure by allowing more clusters.\\

While the EM algorithm achieves a (locally) optimal classification by maximizing the expected likelihood function, this criterion cannot be used to choose the optimal number of clusters: the likelihood is non-decreasing in $K$, thus maximizing the likelihood with respect to $K$ will always give $K \equiv N$; that is each time series is its own class. For parametric models one can use the BIC to choose $K$ \citep{BiernackiGovaert98_choosingmodels}, but for non-parametric settings this is not directly applicable. A common heuristic is the ``elbow rule'', where the number of clusters is determined by looking where the likelihood does not show a substantial increase anymore. 

\citet{CeleuxSoromenho96_EntropyCriterion} propose an entropy based criterion to assess the optimal number of clusters. The \emph{normalized entropy criterion} (NEC) chooses that $K$ which minimizes
\begin{equation}
\label{eq:NEC}
NEC(K) = \frac{E(K)}{\ell^*(K) - \ell(1)}, K \geq 2,
\end{equation}
where $\ell^*(K)$ is the log-likelihood of the best $K$ cluster solution, and 
\begin{equation}
\label{eq:classification_entropy}
E(K) = - \sum_{k=1}^{K} \sum_{i=1}^{N} \gamma_{ik} \log \gamma_{ik} \geq 0, 
\end{equation}
is the entropy of the soft classification matrix. Since it is only based on the class probabilities and the log-likelihood, it can be easily computed even for non-parametric classification methods.

The entropy in \eqref{eq:classification_entropy} measures how well the best $K$ cluster partition can separate the data. In the case of perfectly separable classification, $\gamma_{ik} = 1$ for one $k$ and $0$ otherwise (for each $i$); in this case $E(K) = 0$. In practice, classification is not perfect, thus in general $E(K) > 0$. Hence it makes sense to choose that $K$ which minimizes $E(K)$ as this is as close as possible to a perfect separation. Since the baseline value of the likelihood changes for each $K$, the entropy is normalized by the optimal log-likelihood for each $K$. The optimal number of clusters is the one that minimizes \eqref{eq:NEC}. See \citet{Biernacki99_ImprovementNEC, CeleuxSoromenho96_EntropyCriterion} for details and simulation results.

It must be noted though that rather than looking at the global minimum, it is more useful to consider all local minima as possible candidates. Only focusing on the global minimum can lead to an under-estimation of the true order $K$. For example, sometimes a $K = 2$ cluster solution gives binary weights to each class - and thus $E(K) = 0$ - but can be far from representing the true number of clusters, as it averages over several clusters in one region of the space. Thus for simulations and applications I use the NEC rule combined with the ``elbow'' rule in the log-likelihood to choose an appropriate $K$.


\section{Simulations}
\label{sec:simulations}

This section shows how the methods perform on simulated data. In particular, I consider $K = 5$ sub-systems consisting of both stationary and non-stationary series: one white noise sequence (flat spectrum), two $AR(1)$ processes with $\phi = 0.5$ and $0.75$ respectively, and two sine waves with frequencies $\omega = 0.1$ and $0.2$ (on the $[0, 0.5]$ scale) corrupted by additive Gaussian noise. For each model I generate $n = 100$ series with $T = 50$ observations each. All series have been scaled to zero mean and unit variance.

\begin{figure}[!t]
\centering
\subfloat[ (left) sample time series from each of the $5$ groups: (1) white noise, (2) $AR(1)$ with $\phi = 0.5$, (3) $AR(1)$ with $\phi = 0.75$, (4) $\sin(1/10 2 \pi t / T)$, (5) $\sin(1/20 2 \pi t / T)$; (right) corresponding spectra.]{\includegraphics[width=.45\textwidth]{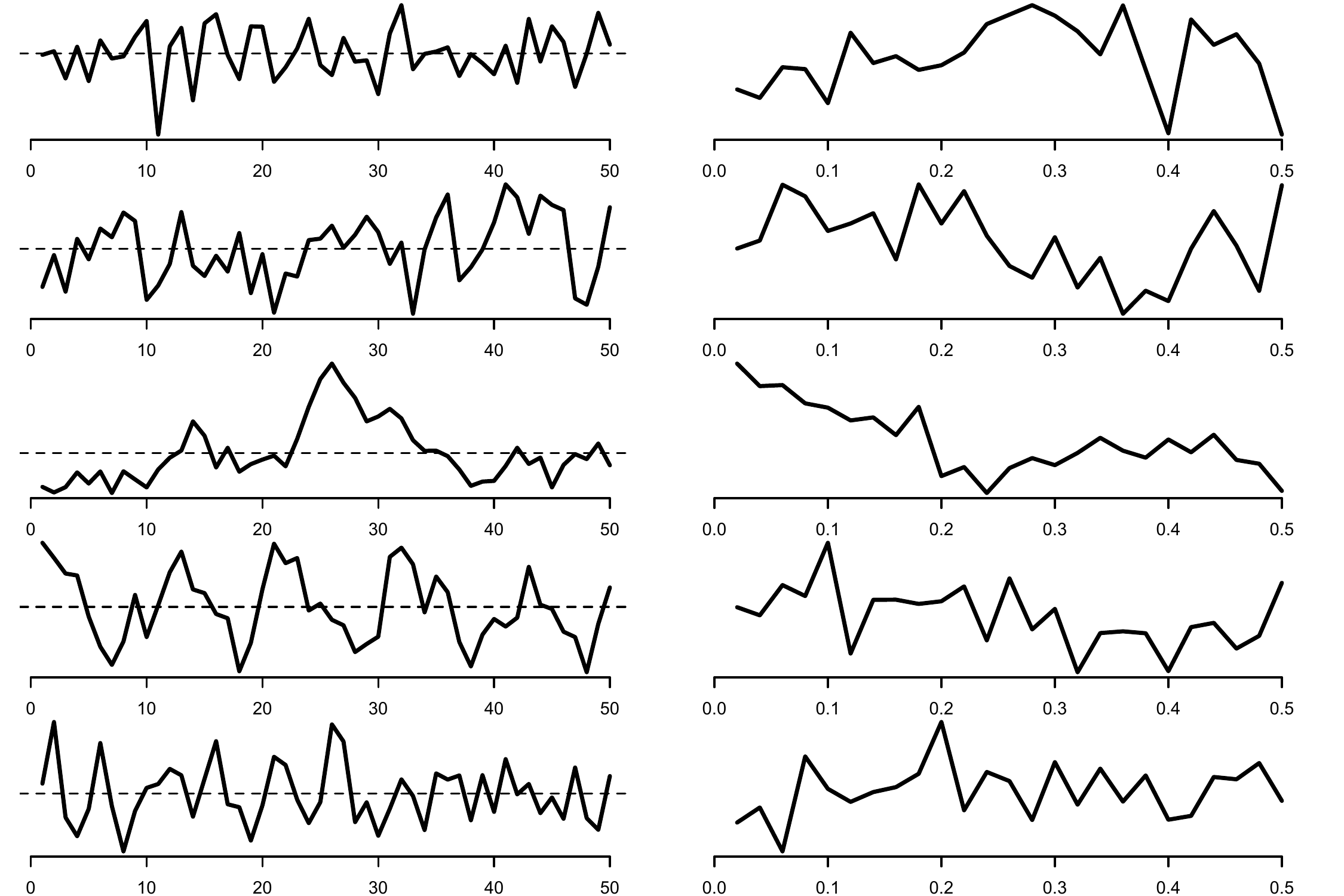}\label{fig:sample_series5} }
\hspace{1cm}
\subfloat[(top) log-likelihood and NEC as a function of $K$; (bottom-left) conditional class probabilities $\gamma_{nk}$; (bottom-right) logarithm of estimated cluster centers equaling optimal estimates $\widehat{f}_{\mathcal{S}_k}(\lambda)$, $k = 1, \ldots, 5$. ]{\includegraphics[width=.45\textwidth]{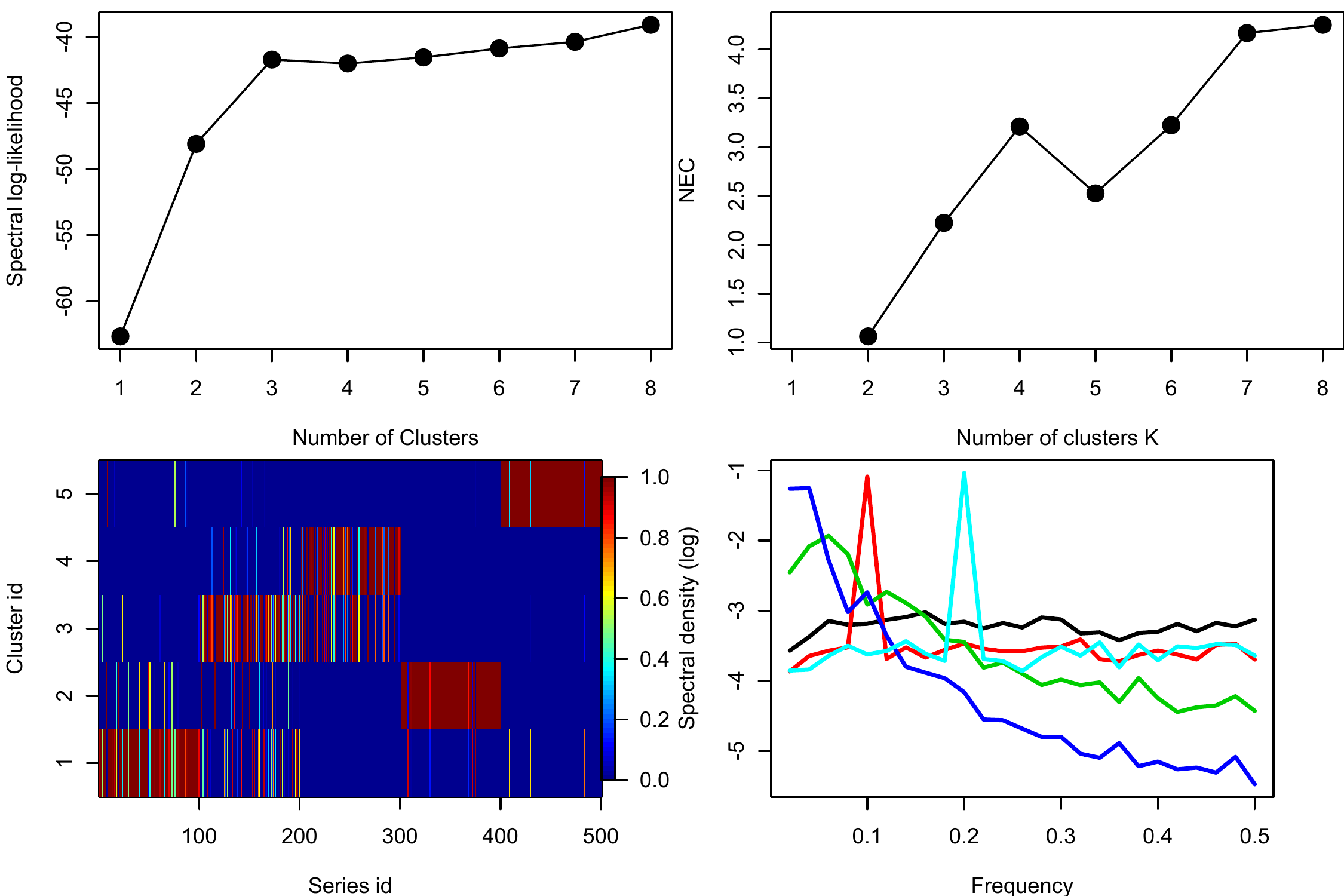}\label{fig:simulation_5series_results} }
\caption{Simulation results for the frequency domain EM algorithm.}
\label{fig:simulation_study}
\end{figure} 

Figure \ref{fig:sample_series5} shows a representative series of each sub-system. All corresponding non-smoothed periodograms have high variance (not consistent estimate). The nonparametric EM can be directly applied to these raw periodograms to cluster the $500$ time series.

The ``elbow'' rule for the log-likelihood favors a $K = 3$ solution, because separating the signals into the two non-stationary signals plus all stationary signals in the third class provides the largest gain in likelihood. The additional likelihood gain by separating the stationary signals into their sub-systems is negligible and thus is not evident in a plot of the log-likelihood as a function of $K$. The NEC has a global minimum at $K = 2$ (one sine wave plus rest) and a local minimum at $K = 5$. The log-likelihood clearly shows that $K = 2$ can not be an optimal separation, thus we take the $K = 5$ local minimum. Figure \ref{fig:simulation_5series_results} shows a very good separation between all signals, except for some cross-matches between the white noise sequence and the two $AR(1)$. However, as the parameters are close to each other and due to the small sample size ($N = 50$), some overlap between them can not be avoided - even using the true model and an MLE $\widehat{\phi}_{MLE}$ to cluster them (see below).

\subsection{Comparison to model based clustering}
\begin{figure}[!t]
\centering
\makebox{\includegraphics[width=0.7\textwidth]{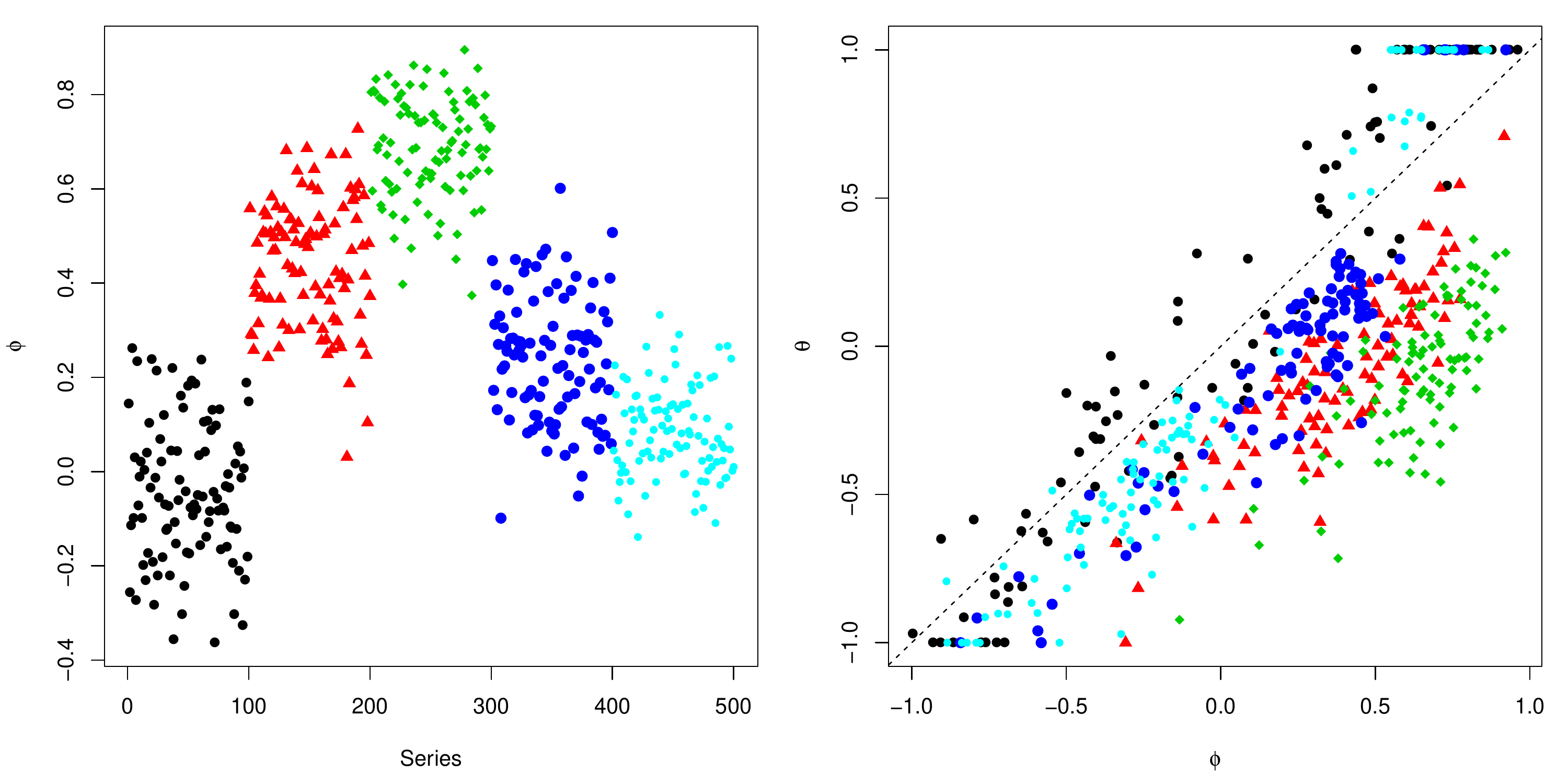}}
\caption{\label{fig:simulation_series5_model_clustering} Model based classification: (left) $\widehat{\phi}$ (y-axis) from fitting an $AR(1)$ to all series; (right) estimate pair $(\widehat{\phi}, \widehat{\theta})$ from fitting an $ARMA(1,1)$ model to each series. Colors and shapes represent the true classes, not estimated clusters from the data.}
\end{figure}

For comparison I also fit $AR(1)$ and $ARMA(1,1)$ models\footnote{The series $x_t$ is an auto-regressive moving average process of order $(1,1)$ if it satisfies $x_t - \phi x_{t-1} = \varepsilon_t - \theta \varepsilon_{t-1}$, where both parameters $\phi$ and $\theta$ must lie in $(-1,1)$ to guarantee stationarity and invertibility.} to each series. Figure \ref{fig:simulation_series5_model_clustering} shows the separation of the series in the parameter space $\phi \in (-1,1)$ and $(\phi, \theta) \in (-1,1) \times (-1,1)$. Using the $AR(1)$ model not only gives a large overlap between the non-stationary signals and the stationary $AR(1)$, $\phi = 0.5$, but also completely fails to distinguish between the two harmonic signals. Even if the true signal is an $AR(1)$, model based clustering still has many falsely classified signals. The overlap in the fitted parameters $\widehat{\phi}_{MLE}$ show that the bad performance of the frequency domain EM for the $AR(1)$ series is not due to the algorithm, but results from the true parameters of distinct $AR(1)$ being very close ($0.5$ and $0.75$). In this case even the maximum likelihood estimator (MLE) provides wrong conclusions. 

Extending the $AR(1)$ to $ARMA(1,1)$ models improves the separability between the two harmonic series, but also leads to additional variation in other regions of the parameter space. In particular, the black dots around the 45-degree line show that avoiding the model bias by simply using a larger model class introduces another problem of model based clustering. Here the model class is an $ARMA(1,1)$, but the true process is white noise, which is a special case of an $ARMA(1,1)$ for $\phi = \theta = 0$. However, every $ARMA(1,1)$ with $\phi = \theta$ also describes a white noise process, thus the MLE finds optimal solutions along the $\phi = \theta$ line and thus adds artificial variance - and thus performance loss for the clustering.\\

The exploratory analysis of the AR and ARMA models is an example of how the model selection bias can undermine clustering algorithms. For a good classification we would need to identify the correct model for each series first, and then estimate the parameters on each tuned model. However, even if we had the time and resources to do a model check for all $N$ time series, the $AR(1)$ example shows that even if we found the true model for each series, a large overlap $\widehat{\phi}_{MLE}$ remains (red triangles and green diamonds in the left panel of Fig. \ref{fig:simulation_series5_model_clustering}). 

The non-parametric EM approach, on the other hand, does not require any modeling and subsequent checks, and has comparable performance to the model based clustering if we knew the true model (white noise and $AR(1)$) and performs much better if the models are wrong (sine waves versus rest).


\section{Applications}
\label{sec:Results}
In this section I demonstrate the usefulness and wide applicability of the presented methods on and income growth (stationary) and neuron spike train (non-stationary) data. 

\subsection{States with similar income dynamics in the US}
\label{sec:marco_econ_results}
\begin{figure}[!t]
\centering
\makebox{\includegraphics[width=0.9\textwidth]{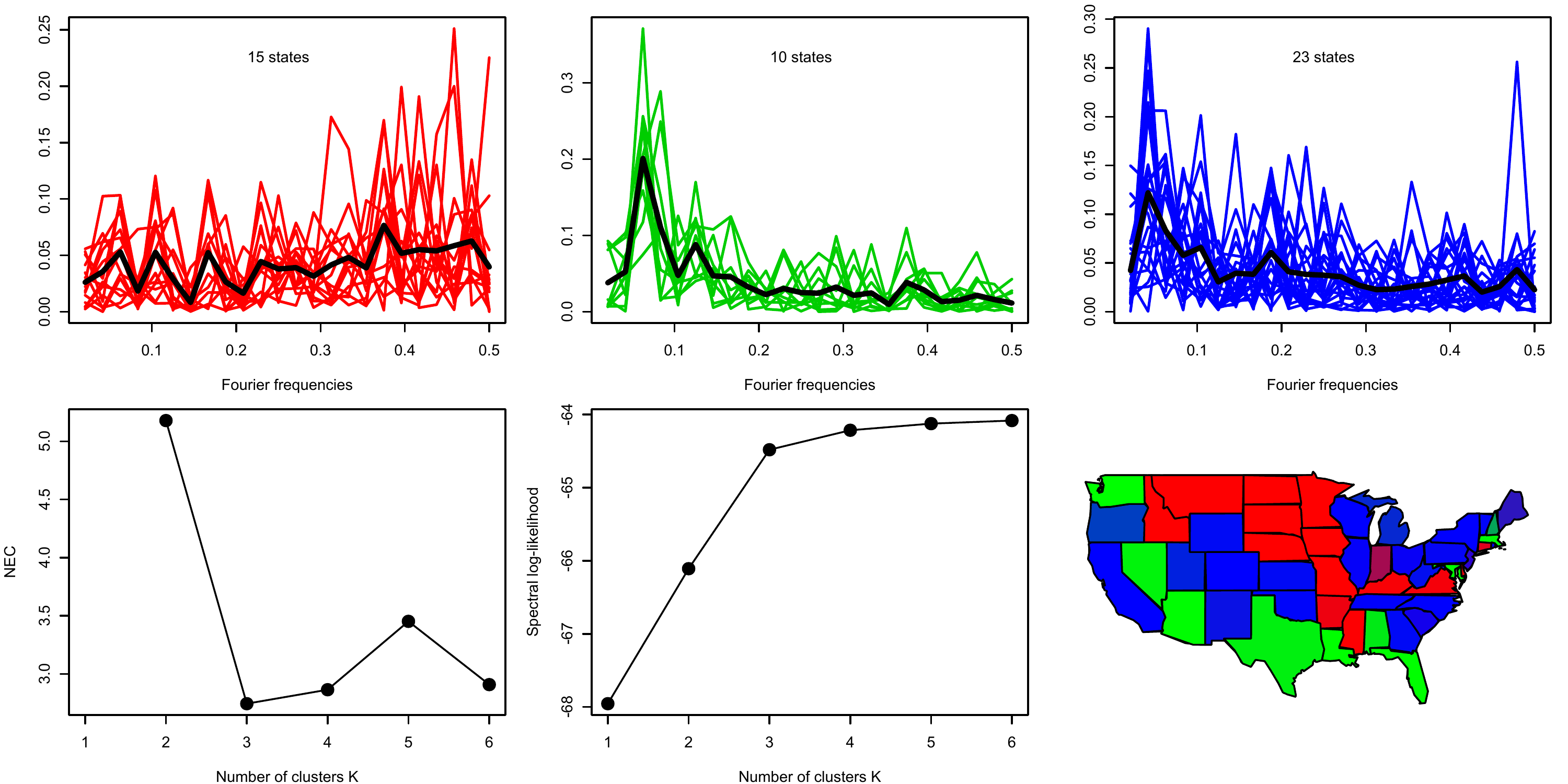}}
\caption{\label{fig:US_growth_1960_2008} Non-parametric, frequency domain EM detects $3$ dominant dynamics of per-capita income growth (top); (left) normalized entropy from \eqref{eq:NEC} as a function of $K$; (center) spectral log-likelihood \eqref{eq:loglik_total} at the optimum for each $K$; (right) color-coded US map where red/green/blue intensity equals the conditional probability $\gamma_{nk}$ (RGB = ($\widehat{\gamma}_{n1},\widehat{\gamma}_{n2}, \widehat{\gamma}_{n3}$)).}
\end{figure}

First, all $48$ (more or less) stationary series $\mathbf{x}_{i,t}$ were transformed via the DFT to get the raw periodograms $I_{\mathbf{x}_{i,t}}(\omega)$, $i=1, \ldots, 48$. Without any further smoothing all $K$-cluster models for $K = 1, \ldots, 6$ were fitted to the data and both the $NEC(K)$ and the log-likelihood suggest that $K = 3$ clusters provide a good fit. The upper row in Fig.\ \ref{fig:US_growth_1960_2008} shows the periodograms of the three classes and the estimate $\widehat{f}_{\mathcal{S}_k}(\lambda)$ (black line) using \eqref{eq:gamma_periodogram}. The x-axis represents the Fourier frequencies $\omega_j$, which have been re-scaled from $[0, \pi]$ to $[0,0.5]$ for easier interpretation. Peaks at frequency $\omega_j$ mean that periods of length $1/\omega_j$ are important for the variation in the data. For example, the blue series show two important low frequencies (long cycles): $\omega \approx 0.04$ and $\omega \approx 0.18$. They correspond to a cycle of $25$ years and $5$-$6$ years -- which represent a generation cycle and a (short) business cycle \citep{Tylecote94_Longcycles}. Note that $AR(1)$ models \citep{Dhiral01_ClusteringARIMA} may be appropriate for the red dynamics ($AR(1)$ coefficient slightly negative), but cannot capture two cycles as shown in the blue and green periodograms.\\ 

The spatial connectivity of the obtained clusters confirms the good model fit as it separates US economy $\mathcal{S}$ in three major sub-economies/regions:\footnote{Any resemblance of the RGB color system to politics is purely coincidental.}
\begin{description}
\item[$\approx$ East \& Rockies \& CA (blue):] economy is highly persistent, changes are slow; business cycle of $\approx 5$-$6$ years is also important.
\item[$\approx$ South-West (green):] also highly persistent and affected by global business cycle of $7-8$ years (peak at $\omega \approx 0.13$).
\item[$\approx$ Mid-West (red):] almost flat spectrum, high frequencies (short cycles) are slightly more important; decoupled from global business cycle.
\end{description}
One possible explanation why the red states have a flat spectrum, is that they are mostly agricultural states, and since people have to eat no matter how the global economy is doing, the red states' income is not affected too much by recession or other market fluctuations. On the contrary, states whose economy - and thus income - relies heavily on industry, production, or technology are more affected by global economy swings, which typically happen every $7$-$8$ years.\\

Hence, the classification map in Fig.\ \ref{fig:US_growth_1960_2008} can provide a basis for more effective policies to boost local economies facing a recession: it might be more effective to allocate main parts of public investments to states that are actually affected by the business cycle, and not put it in states which are decoupled from global economy.

\subsection{Spike sorting}
\label{sec:features}
For the neuron classification we can either try to fit a mixture model directly on the $T$ - dimensional data, or compute ``features'' for each spike that summarize its shape. A good feature selection will reduce the dimensionality of the data, and thus greatly accelerate computations. 

Here I will cluster both in the time and frequency domain: for the first I fit a Gaussian mixture model (GMM) on the logarithm of the slowness of each spike, $\log \Delta \left( s_{j,t} \right)$; for the second I use the frequency domain EM algorithm on the power spectra induced by each spike $s_{j,t}$.

\subsubsection{Gaussian mixture model on slowness}
The histogram in Fig.\ \ref{fig:extracted_spikes} of $\lbrace \log \Delta \left( s_{j,t} \right) \rbrace_{j=1}^{1,747}$ shows 5-6 peaks, which presumably correspond to 5-6 differently shaped spikes. Thus I fit GMMs to $\log \Delta \left( s_{j,t} \right)$ and assign each spike $s_{j,t}$ to the cluster with highest a posteriori probability. Table \ref{tab:gaussian.mix} shows parameter estimates of the $6$ component model, which was chosen according to the highest BIC score (Fig.\ \ref{fig:extracted_spikes}) from all GMMs up to order $10$.\footnote{To avoid local maxima, I ran the EM algorithm (package \texttt{mixtools} in R) $100$ times for each $K$ and chose the largest local optimum solution in each run.}

\begin{table}[!t]
\begin{center}
\caption{\label{tab:gaussian.mix} EM estimates of a $6$ component GMM for $\log \Delta(s_{j,t})$}
\begin{tabular}{rrrrrrr}
  \hline
 & Comp 1 & Comp 2 & Comp 3 & Comp 4 & Comp 5 & Comp 6 \\ 
 \hline
 \hline
$\pi_k$ & 0.069 & 0.218 & 0.093 & 0.511 & 0.078 & 0.031 \\
  \hline
$\mu$ & -3.285 & -2.766 & -2.331 & -2.171 & -1.671 & -1.442 \\ 
$\sigma^2$ & 0.155 & 0.171 & 0.037 & 0.156 & 0.125 & 0.042 \\ 
   \hline
\end{tabular}
\end{center}
\end{table}

The corresponding spikes are shown in the upper right panel of Fig.\ \ref{fig:extracted_spikes}. As $tol = 0.25$ was too conservative, two shapes still represent noise, and GMM identifies $K=4$ different neurons.

\subsubsection{Clustering in the frequency domain}
After time-domain techniques, I use the frequency domain EM algorithm described in Section \ref{sec:EM_spectral_density}. An additional advantage of working in the frequency domain compared to the time-domain is that misalignment of the spikes does not affect the clustering. 

\begin{figure}[!t]
\centering
\makebox{\includegraphics[width=0.75\textwidth]{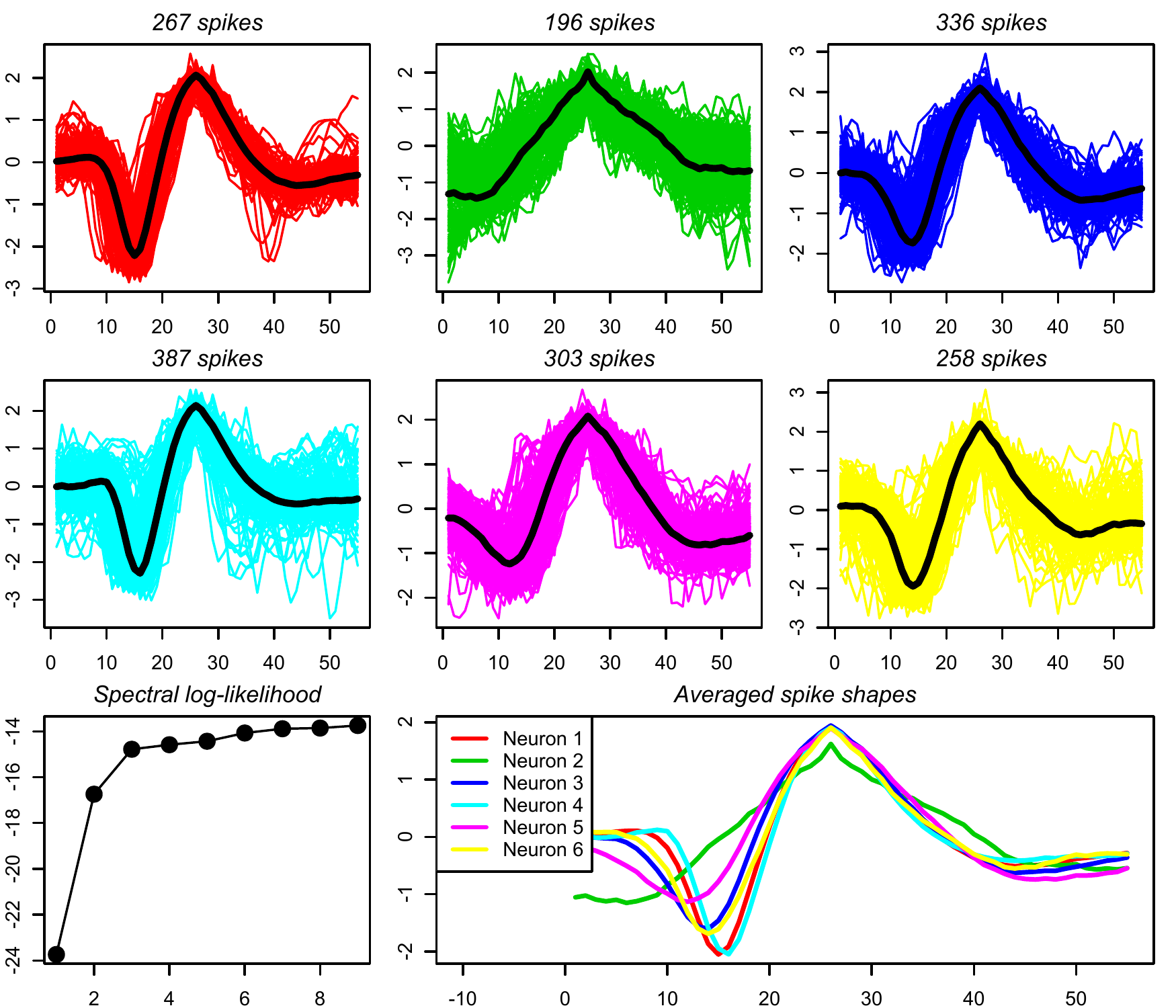}}
\caption{\label{fig:EM_spikes_periodogram} EM on periodograms of spike signals $s_{j,t}$ with $K = 6$ clusters.}
\end{figure}

Also here I fit all mixture models up to order $K = 9$. In this case $NEC(K)$ achieves a global minimum at $K = 2$ and is monotonically increasing (not shown here). However, the two cluster solution is only optimal in the sense that it separates perfectly between spikes and noise, even though there is a relevant sub-classification within all spikes (similar to the behavior of $NEC(K)$ in the simulations). Hence here I use the ``elbow'' rule in the log-likelihood to determine the number of clusters. The most prominent ``elbow'' occurs at $K = 3$ (Fig.\ \ref{fig:EM_spikes_periodogram}) followed by another level-shift at $K = 6$. Since $K = 6$ was also by the BIC for the time-domain classification, I choose $K = 6$ for easier comparison. The $K = 6$ cluster solution reveals five spikes and one noise class (green shapes). Thus compared to the time-domain technique the frequency domain version detects one more spike.

\section{Discussion and outlook}
I introduce a novel technique to detect and classify similar dynamics in signals, where similar dynamics can either mean similar shape for non-stationary signals, or similar auto-correlation for stationary signals. It is an adaptation of the EM algorithm to the power spectra of the signals and thus future research can benefit from the extensive literature in both areas of signal processing. Applications to neural spike sorting and pattern recognition in macro-economic time series demonstrate the usefulness of the presented method.

I also used the recently introduced slowness feature for the classification of neuron spikes. The slowness of signals can separate signals from noise and also distinguish differently shaped signals. Compared to multivariate methods in the literature it is very fast and easily computable, and more robust to outliers than for example the standard approach of a simple threshold method.


\newpage
\phantomsection
\addcontentsline{toc}{section}{References}
\bibliographystyle{chicago}	
\bibliography{NonparametricFrequencyDomainEM}

\newpage
\appendix
\section{Data}
\label{sec:Data}
\paragraph{Spikes:} The \texttt{PKdata} data set can be obtained from \url{www.biomedicale.univ-paris5.fr/physcerv/C_Pouzat/Data.html}. It contains recordings of the electro-chemical signal in the cerebral slice of a rat. A band pass filter for frequencies between $300$ Hz and $5$ kHz has been applied to the signal $y_t$, which was sampled at a rate of $15$ Khz for $1$ minute. 

\paragraph{Income:} The dataset can be obtained from \url{www.bea.gov/regional/spi}. It contains yearly ($1958 - 2008$) average per-capita income of the ``lower $48$'' and the entire US: $I_{j,t}$, $j = 1, \ldots, 49$. As $I_{j,t}$ grew exponentially over time, one typically considers income growth rates $r_{j,t} = \log I_{j,t} - \log I_{j,t-1}$ - also known as log-returns - which are (more or less) stationary. Since we are interested in the individual dynamics of a state compared to the US, I analyze the difference between each state's growth rate to the US, as this is a more refined indicator of the state's dynamics (it removes the overall seasonal dynamics of the US baseline). 

\section{KL divergence and maximum likelihood}
\label{sec:KL_MLE}
Let $p_k := \Prob(X = a_k)$ define a probability distribution for the RV $X$ taking values in the finite alphabet $\mathcal{A} := \lbrace a_1, \ldots, a_K \rbrace$. The Kullback-Leibler (KL) divergence between two discrete probability distributions $p = \lbrace p_1, \ldots, p_K \rbrace$ and $q = \lbrace q_1, \ldots, q_K \rbrace$ 
\begin{equation}
\KL{p}{q} := \sum_{i=1}^{K} p_i \logtwo \frac{p_i}{q_i}  = \E_p \logtwo \frac{p_i}{q_i}
\end{equation}
measures how far $p$ is from the ``truth'' $q$; in particular, if $p = q$ then $\KL{p}{q} = 0$. 

Let $\tilde{p}(x)$ be the empirical distribution function (edf) of a sample $\mathbf{x} = (x_1, \ldots, x_N)$
\begin{equation}
\label{eq:empirical_distribution}
\tilde{p}(\mathbf{x}) := \frac{1}{N} \sum_{n=1}^{N} \delta(x - x_n),
\end{equation}
where $\delta(y)$ is the Dirac delta function, and let $p(\mathbf{x} \mid \theta)$ be a model (distribution) for the RV $X$. The maximum likelihood estimator (MLE) is that $\theta$ which maximizes the log-likelihood of the data (assuming iid)
\begin{equation}
\label{eq:loglik}
\ell( \theta \mid \mathbf{x}) = \sum_{n=1}^{N} \log p(x_n \mid \theta).
\end{equation}
In terms of the KL divergence it is intuitive to select that $\theta$ which minimizes the distance between the empirical distribution of the data, $\tilde{p}(\mathbf{x})$, and the model $p(\mathbf{x} \mid \theta)$. In fact, they are equivalent since 
\begin{eqnarray}
\label{eq:KL_entropy_likelihood}
\KL{\tilde{p}(\mathbf{x})}{p(\mathbf{x} \mid \theta)} = \int \tilde{p}(\mathbf{x}) \log \frac{\tilde{p}(\mathbf{x})}{p(\mathbf{x} \mid \theta)} d \mathbf{x} = - H(\tilde{p}(\mathbf{x})) - \int \tilde{p}(\mathbf{x}) \log p(\mathbf{x} \mid \theta) d \mathbf{x},
\end{eqnarray}
where $H(\tilde{p}(\mathbf{x})) =  - \int \tilde{p}(\mathbf{x}) \log \tilde{p}(\mathbf{x}) d \mathbf{x} $ is the entropy of $\tilde{p}(\mathbf{x})$, which is independent of $\theta$. Thus
\begin{equation}
\label{eq:KL_equiv_loglik}
\arg \min_{\theta} \KL{\tilde{p}(\mathbf{x})}{p(\mathbf{x} \mid \theta)} = \arg \max_{\theta} \E_{\tilde{p}} \log p(\mathbf{x} \mid \theta).
\end{equation}

Plugging \eqref{eq:empirical_distribution} in the right hand side of \eqref{eq:KL_equiv_loglik} shows the equivalence of KL divergence minimization and log-likelihood maximization as
\begin{eqnarray}
\E_{\tilde{p}} \log p(\mathbf{x} \mid \theta) &=& \frac{1}{N} \int \sum_{n=1}^{N} \delta(x - x_n) \log p(x \mid \theta) dx = \frac{1}{N} \sum_{n=1}^{N} \log p(x_n \mid \theta) \\
\label{eq:1_N_loglik}
&=& \frac{1}{N} \ell( \theta \mid \mathbf{x}).
\end{eqnarray}

Conversely the log-likelihood of $\mathbf{x}$ can be computed by 
\begin{eqnarray}
\label{eq:loglik_equals_KL_ent}
\ell( \theta \mid \mathbf{x}) = -N \cdot \left[ \KL{\tilde{p}(\mathbf{x})}{p(\mathbf{x} \mid \theta)} + H(\tilde{p}(\mathbf{x})) \right],
\end{eqnarray}
and consequently
\begin{equation}
\label{eq:probability_from_loglik}
\Prob \left( \mathbf{x} \mid \theta \right) = e^{\ell( \theta \mid \mathbf{x})}.
\end{equation}
Equations \eqref{eq:loglik_equals_KL_ent} and \eqref{eq:probability_from_loglik}  play a key role in the non-parametric EM algorithm defined on the power spectra, as they allow to compute $\ell( \theta \mid \mathbf{x})$ even though $\mathbf{x}$ has not been observed directly, but just its edf $\tilde{p}(\mathbf{x})$ and a model $p(\mathbf{x} \mid \theta)$. In this frequency domain framework, the data $\mathbf{x}$ are the unobserved Fourier frequencies $\omega_0, \ldots, \omega_{T-1}$, the edf $\tilde{p}(\mathbf{x})$ is the periodogram $I_{T,x}(\omega_k)$, and the ``true'' model $p(\mathbf{x} \mid \theta)$ is the EM estimate $\widehat{f}_{\mathcal{S}_k}(\lambda) \mid_{\lambda = \omega_j}$ of the spectral density of sub-system $\mathcal{S}_k$ - see \eqref{eq:est_f_j2}. Thus the conditional probability $\gamma_{ik} = \Prob(z_{ik} = 1 \mid \mathbf{x}_{i,t})$ can be computed by
\begin{equation}
\gamma_{ik} = \frac{e^{\ell \left( I_{\mathbf{x}_{i,t}}(\lambda) \mid \widehat{f}_{\mathcal{S}_k}(\omega) \right) }}{\sum_{k=1}^{K} e^{\ell \left( I_{\mathbf{x}_{i,t}}(\lambda) \mid \widehat{f}_{\mathcal{S}_k}(\omega) \right) }}, \quad i =1, \ldots, N \text{ and } k = 1, \ldots, K.
\end{equation}


\end{document}